\documentclass[runningheads]{llncs}
\usepackage{amsmath,multirow,graphicx,color}
\usepackage[normalem]{ulem}

\useunder{\uline}{\ul}{}
\begin{document}
\title{Depth-agnostic Single Image Dehazing}
%
%

\author{Honglei Xu\inst{1} \and Yan Shu\inst{1} \and
Shaohui Liu\inst{1}}

\institute{Harbin Institute of Technology}
%
\maketitle              
\begin{abstract}
Single image dehazing is a challenging ill-posed problem. Existing datasets for training deep learning-based methods can be generated by hand-crafted or synthetic schemes. However, the former often suffers from small scales, while the latter forces models to learn scene depth instead of haze distribution, decreasing their dehazing ability. To overcome the problem, we propose a simple yet novel synthetic method to decouple the relationship between haze density and scene depth, by which a depth-agnostic dataset (DA-HAZE) is generated. Meanwhile, a Global Shuffle Strategy (GSS) is proposed for generating differently scaled datasets, thereby enhancing the generalization ability of the model. Extensive experiments indicate that models trained on DA-HAZE achieve significant improvements on real-world benchmarks,  with less discrepancy between SOTS and DA-SOTS (the test set of DA-HAZE). Additionally, Depth-agnostic dehazing is a more complicated task because of the lack of depth prior. Therefore, an efficient architecture with stronger feature modeling ability and fewer computational costs is necessary. We revisit the U-Net-based architectures for dehazing, in which dedicatedly designed blocks are incorporated. However, the performances of blocks are constrained by limited feature fusion methods. To this end, we propose a Convolutional Skip Connection (CSC) module, allowing vanilla feature fusion methods to achieve promising results with minimal costs. Extensive experimental results demonstrate that current state-of-the-art methods equipped with CSC can achieve better performance and reasonable computational expense, whether the haze distribution is relevant to the scene depth.

\end{abstract}
\section{Introduction}
\label{sec:intro}
Haze  is a common atmospheric phenomenon. Haze images usually suffer from noticeable visual quality degradation in contrast or color distortion \cite{tan2008visibility} impacting the reliability of models in high-level vision tasks. To improve the overall scene visibility, many image dehazing methods have been proposed to recover the latent haze-free image from the single hazy input.

For the training of image dehazing models, preparing a suitable dataset is the first step. By utilizing professional haze machines, real haze images can be generated \cite{ancuti2018hazeo,ancuti2018hazei,ancuti2016d,ancuti2020nh}. However, the
primary issue of these hand-crafted real datasets is their small-scale data because it is very difficult to collect the haze image and haze-free image at the same scene. Therefore, models trained on them have limited generalization. To solve this problem, Atmospheric Scattering Model (ASM) \cite{narasimhan2003contrast} is applied to synthesize large-scale datasets,
which have boosted the performances of many image dehazing methods. 

\begin{figure}[h]
    \begin{minipage}[b]{.3\linewidth}
        \centering
        \includegraphics[width=1\linewidth]{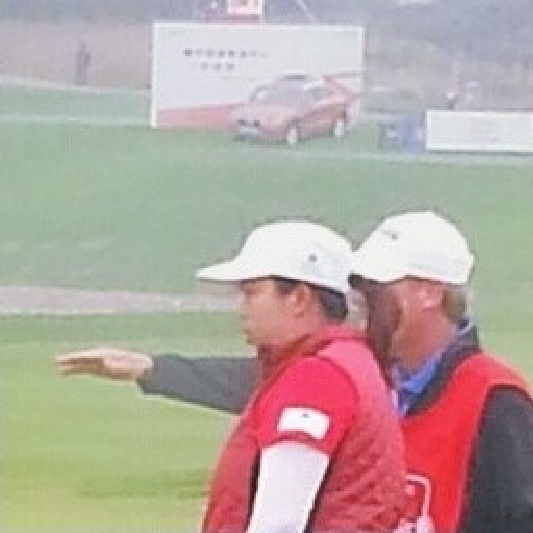}
        \centerline{(a) Hazy}\medskip
    \end{minipage}
    \hfill
    \begin{minipage}[b]{.3\linewidth}
        \centering
        \includegraphics[width=1\linewidth]{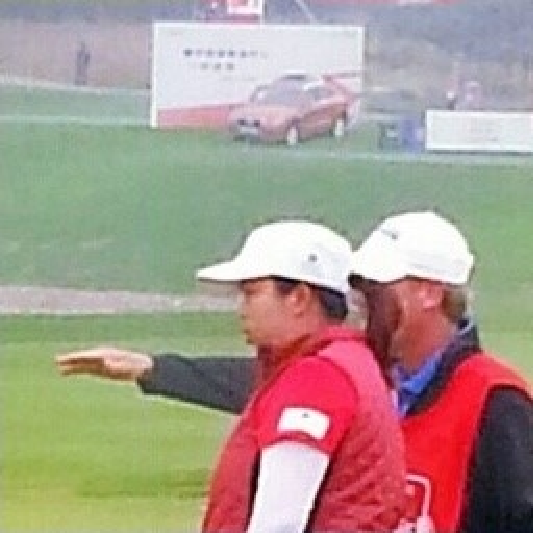}
        \centerline{(b) Restored}\medskip
    \end{minipage}
    \hfill
    \begin{minipage}[b]{.3\linewidth}
        \centering
        \includegraphics[width=1\linewidth]{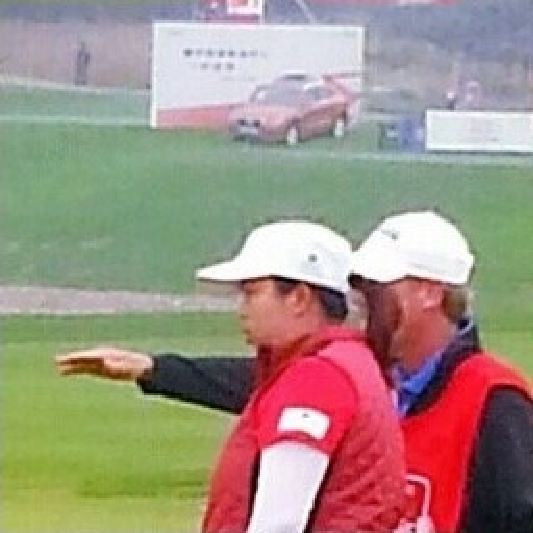}
        \centerline{(c) GT}\medskip
    \end{minipage}
    
    \begin{minipage}[b]{0.3\linewidth}
        \centering
        \includegraphics[width=1\linewidth]{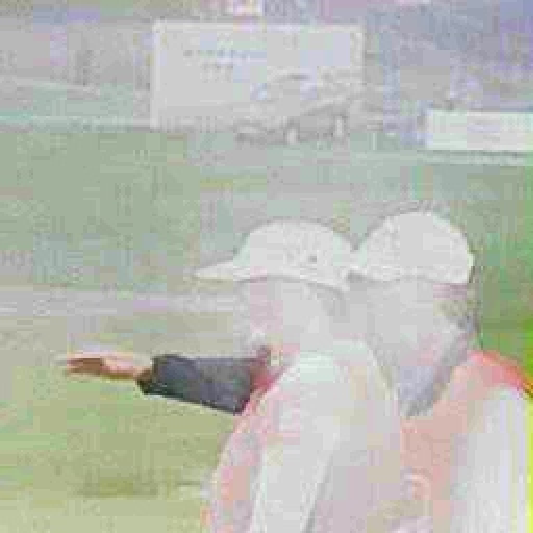}
        \centerline{(d) Difference}\medskip
    \end{minipage}
    \hfill
    \begin{minipage}[b]{.3\linewidth}
        \centering
        \includegraphics[width=1\linewidth]{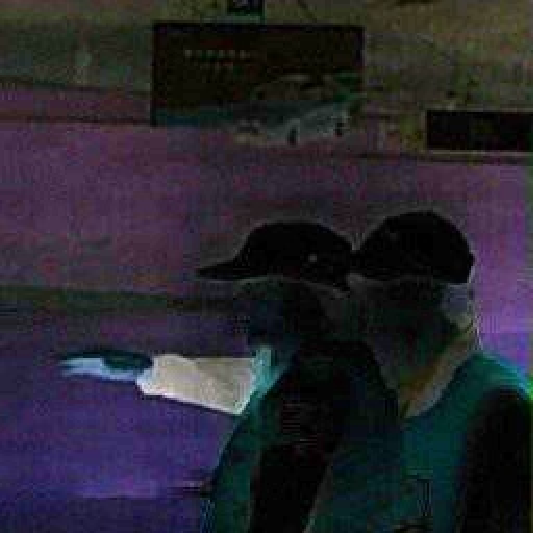}
        \centerline{(e) Absolute of difference}\medskip
    \end{minipage}
    \hfill
    \begin{minipage}[b]{.3\linewidth}
        \centering
        \includegraphics[width=1\linewidth]{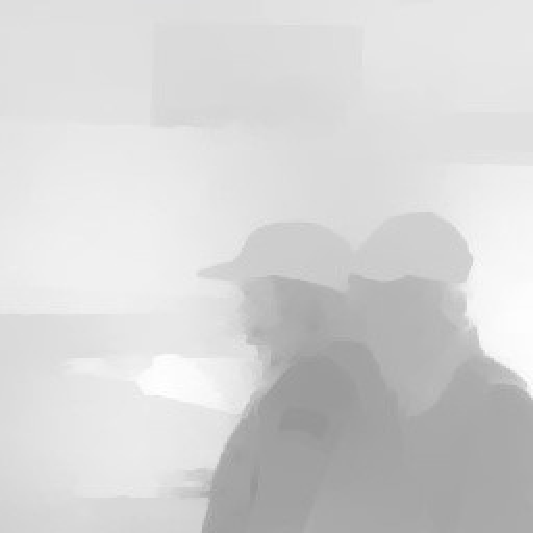}
        \centerline{(f) Depth}\medskip
    \end{minipage}
    \caption{(a) is the hazy image. (b) is the restored image generated by the model trained on OTS. (c) is the GroundTruth. (d,e) shows the difference between (b) and (c). (f) is the depth map to stnthesize (a).}
    \label{error}
\end{figure}

\begin{figure}[h]
    \begin{minipage}[b]{.594\linewidth}
        \centering
        \includegraphics[width=1\textwidth]{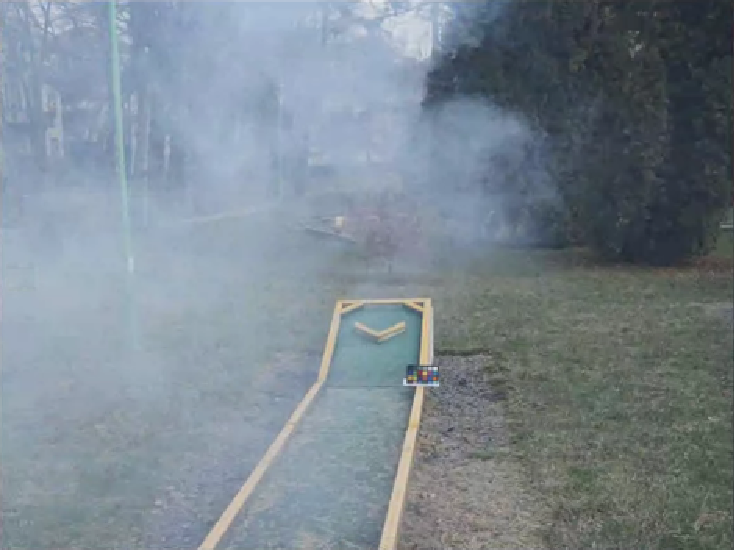}
        \centerline{(a)}\medskip
    \end{minipage}
    \hfill
    \begin{minipage}[b]{.396\linewidth}
        \centering
        \includegraphics[width=1\textwidth]{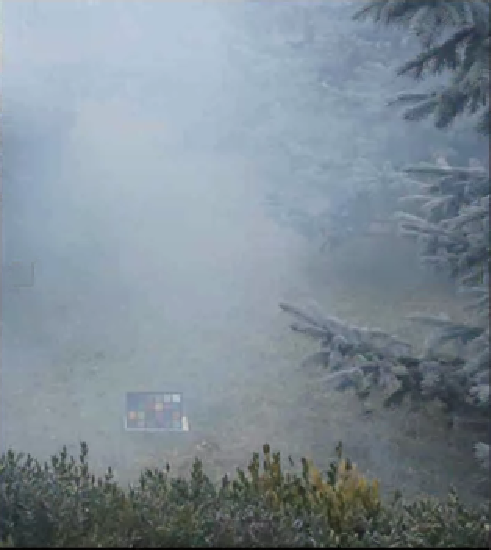}
        \centerline{(b)}\medskip
    \end{minipage}
    \caption{(a) Non-homogeneous hazy image. (b) Homogeneous hazy image.}
    \label{himg}
\end{figure}

Despite this, challenges still exist, particularly in real-world dehazing tasks, existing state-of-the-art image dehazing method trained on synthetic large-scale datasets achieves limited performances in the real-world dehazing task. Due to the inherent characteristics of synthetic methods, where the haze density and scene depth are positively correlated, these priors is not absolute in the real word, as shown in Figure \ref{himg} (a). Moreover, they may mislead models to take shortcuts by relying on scene depth instead of learning the essence of haze distribution. As shown in Figure \ref{error}, the quality of the restored region will be low if the haze density does not comply with the depth prior.

In addition, Depth-agnostic dehazing is a more complicated task because of the lack of depth prior. Therefore, an efficient and effective architecture is necessary. Existing dehazing methods based on U-Net architecture have designed dedicated blocks to improve performances. In consideration of performances and computational expenses, vanilla feature fusion methods (adding and adding-like operation) have been applied to reduce the input dimension, decreasing the optimal performances of the blocks. These two intrinsic defects hinder the practical applications of existing dehazing models. 

In this paper, we try our best to answer two questions. \textbf{ 1) Can we generate a large-scale dataset where
the haze distribution is independent of scene depth? 2)  Is it possible to unleash the full potential of specialized blocks even with vanilla feature fusion methods?} To this end, we propose Depth-agnostic Single Image Dehazing.
 
\begin{figure}
    \begin{minipage}[b]{1.0\linewidth}
      \centering
      \centerline{\includegraphics[width=\textwidth]{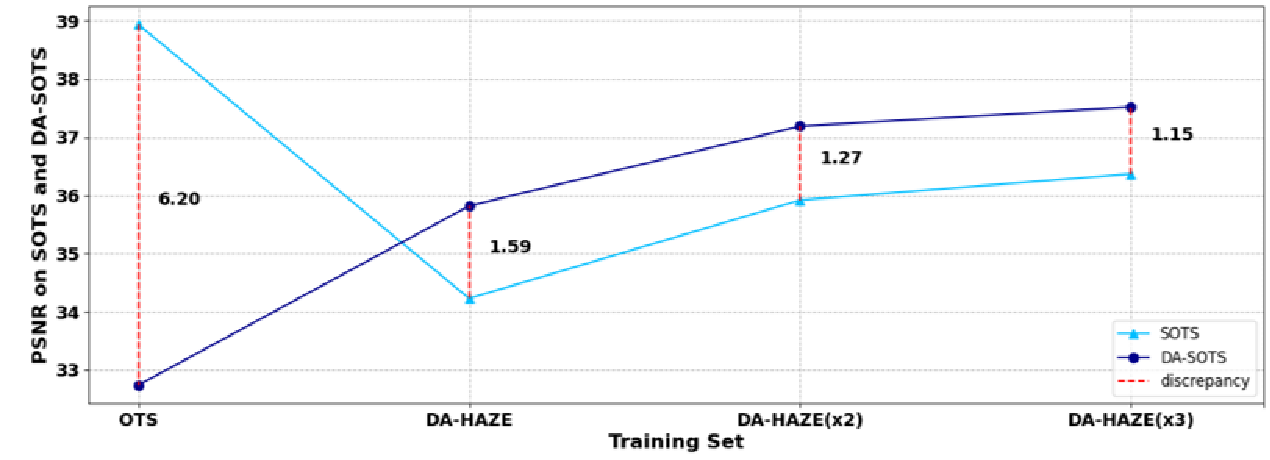}}
    \end{minipage}
    
    \begin{minipage}[b]{.5\linewidth}
      \centering
      \centerline{\includegraphics[width=\textwidth]{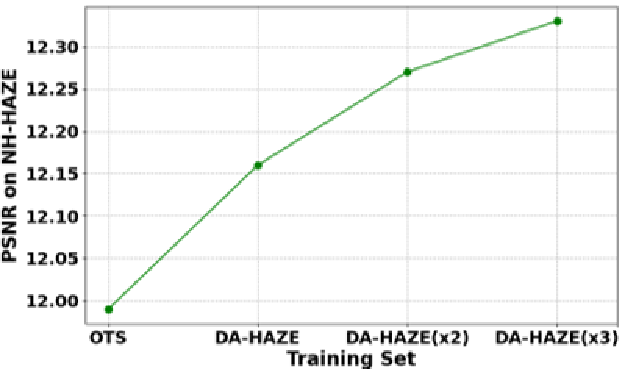}}
    \end{minipage}
    \begin{minipage}[b]{.5\linewidth}
      \centering
      \centerline{\includegraphics[width=\textwidth]{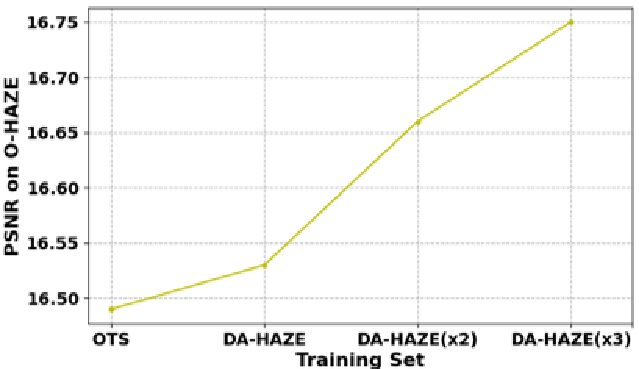}}
    \end{minipage}
    \caption{Comparison of the results generated by  NAF-Net trained on the OTS and our DA-HAZE and tested on the corresponding test sets SOTS and DA-SOTS (the top figure), real-world dataset NH-HAZE and O-HAZE (bottom figures). DA-HAZE($\times$n) denotes various scales of the dataset.}
    \label{compare}
\end{figure}

On the one hand, we propose a novel synthetic method to generate a depth-agnostic dataset, named DA-HAZE, in which the haze density is not correlated with the scene depth. Specifically, by leveraging ASM, we generate sufficient data in DA-HAZE to enable large-scale training. Moreover, the pairs between haze-free images and depth maps are randomly ordered. This allows the models to perceive the haze distribution instead of being misled by scene depth. We also propose a novel Global Shuffle Strategy (GSS) to make DA-HAZE scalable, in which the same image matches multiple depth estimations so that the generalization of models can be improved. On the other hand, we propose a Convolutional Skip Connection (CSC) module, by introducing a normal convolution layer to the connection path, vanilla feature fusion methods such as adding or adding-like operations can achieve promising results, with minimal computational costs. 

The contributions of this work are threefold:

(i) We propose a novel synthetic method to generate a depth-agnostic dataset (DA-HAZE) where the relationship between haze density and scene depth is decoupled. This guides models to  perceive the haze distribution instead of relying on scene depth.

(ii) We propose a Global Shuffle Strategy (GSS) for making DA-HAZE scalable, which can improve the generalization ability of models. Experiments show that models trained on DA-HAZE with GSS achieve significant improvements on real-world benchmarks compared to the previous OTS dataset, with less discrepancy on different distributed validation sets as illustrated in Figure \ref{compare}.

(iii) We propose a Convolutional Skip Connection (CSC) module, allowing for vanilla feature fusion methods to achieve promising results with minimal costs. Experiments show CSC brings a significant gain to the current dehazing methods whether haze distribution is relevant to scene depth, with a promising trade-off between performance and computational expenses. 

\section{Related Work}
\subsection{Image Dehazing datasets}

\textbf{Hand-crafted real datasets} generate haze in scene images by specific haze machines. \cite{ancuti2018hazeo} firstly introduced a database termed O-HAZE in outdoor scenes. Considering that haze is not uniformly distributed in many cases, NH-HAZE \cite{ancuti2020nh}, a non-homogeneous realistic dataset with real hazy and corresponding haze-free images was proposed. Generally, these datasets share similar distributions to the real-world data, however, their limited scales hinder better generalization of the models trained on them.

\textbf{ASM synthetic datasets} consider that it is difficult to collect the haze image and haze-free image at the same scene. Thus, the atmospheric scattering model (ASM) is utilized to generate large-scale synthetic data. The most commonly applied dataset is the Realistic Single Image Dehazing (RESIDE) \cite{li2018benchmarking}, which consists of Indoor Training Set, Outdoor Training Set, and Synthetic Objective Testing Set. Most of the methods trained on the benchmark achieved promising results, but suffer from limited performance towards real-world images due to the significant domain gaps.

Our proposed depth-agnostic dataset (DA-HAZE) inherits the merits of handcrafted real datasets and ASM synthetic datasets. On the one hand, we ensure its large scale based on ASM, on the other hand, we modify the synthetic methods to generate depth-agnostic hazy images.

\subsection{Single Image Dehazing}
Single image dehazing methods \cite{he2010single,zhu2015fast,cai2016dehazenet,ren2016single,li2017aod,qin2020ffa,xu2023aprogressive} can be divided into two categories (Prior-based and  Data-driven methods).  We also introduce the classical image dehazing methods based on U-Net architecture relevant to our works.


\textbf{Prior-based methods} rely on the ASM and hand-crafted priors, such as dark channel prior (DCP) \cite{he2010single} and color attenuation prior (CAP) \cite{zhu2015fast}. He et al. proposed DCP based on a key observation - most local patches in haze-free outdoor images contain some pixels with very low intensities in at least one color channel, which can help estimate the transmission map. CAP established a linear relationship between depth and the difference between brightness and saturation. These methods only work well in specific scenes which happen to satisfy their assumptions.

\textbf{Data-driven methods} aim to directly or indirectly learn the mapping function. DehazeNet \cite{cai2016dehazenet} and MSCNN \cite{ren2016single} utilized CNNs to estimate the transmission map. AOD-Net \cite{li2017aod} rewrote the ASM and estimated atmospheric light together with the transmission map. However, the cumulative errors introduced by inaccurate estimations of the transmission map and atmospheric light may cause performance degradation. To avoid this, recent works tend to recover the haze-free image from the hazy image directly by deep neural networks. To enhance the feature fusion, MSBDN proposed a boosting strategy and back-projection technique. FFA-Net \cite{qin2020ffa} dealt with different types of information by introducing the feature attention mechanism. A novel contrastive regularization was proposed in AECR-Net \cite{wu2021contrastive} so that it can benefit from both positive and negative samples. Although these methods significantly develop the dehazing performances, the complexity of networks also increases. 

\textbf{U-Net Architecture} was originally proposed for image segmentation, which is similar to an autoencoder but with skip connections between the encoder and decoder to preserve fine-grained details. Many researchers have extended this framework into image dehazing. Zhao et al. designed the HyLoG-ViT \cite{zhao2021hybrid} to capture local and global dependencies, with promising results achieved but suffering from large model complexity. To achieve a tradeoff between performance and computational cost, Dehazeformer \cite{song2023vision} and DEA-Net \cite{chen2023dea}  applied adding-like methods
by which the input dimension to the block can be reduced. Nevertheless, these designs hinder the full exploration of the proposed blocks. Distinguished from theirs, we revisit the U-Net Architecture and propose a Convolutional Skip Connection (CSC) method to address the dilemma.

\section{Method}
\subsection{Depth-agnostic Dataset  }

\textbf{Genaration Method.} In previous methods for generating dehazing datasets, ASM (Atmospheric Scattering Model) has been utilized to synthesize haze data. The formulation of this process is as follows:
\begin{equation}
\label{asm}
    I=J \cdot t+A \cdot(1-t),
\end{equation}
where $I$ is the observed hazy image, $J$ is the underlying haze-free image to be recovered, $A$ is the global atmospheric light, indicating the ambient light intensity and $t$ is the transmission map, which represents the distance-dependent factor affecting the fraction of light that reaches the camera sensor and is expressed as:
\begin{equation}
\label{trans}
    t = e^{- \beta d}
\end{equation}
where $\beta$ is the scattering coefficient of the atmosphere, and $d$ is the depth map. In this case, there is a positive correlation between the haze distribution and scene depth. However, models can be misled by this prior information and may disregard the haze density information.

\begin{figure}[h]
    \begin{minipage}[b]{.329\linewidth}
		\centering
		\includegraphics[width=1\linewidth]{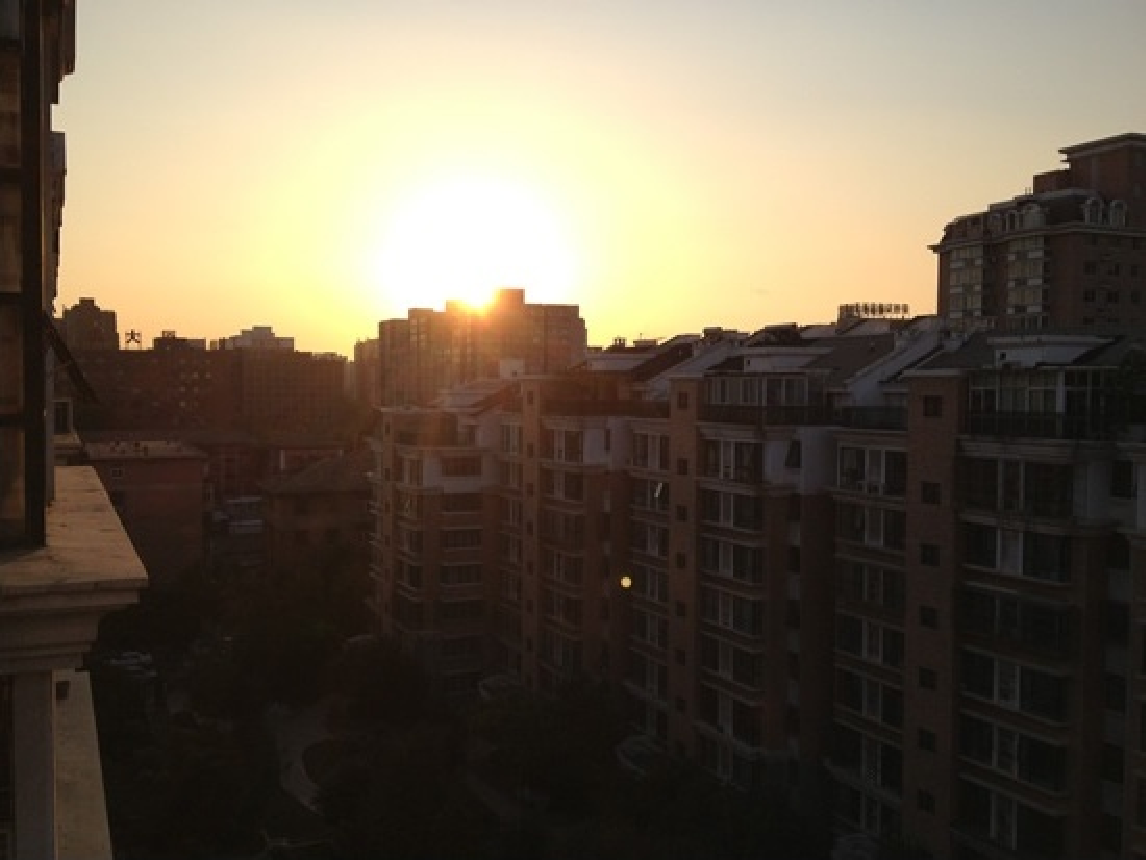}
	\end{minipage}
	\begin{minipage}[b]{.329\linewidth}
		\centering
		\includegraphics[width=1\linewidth]{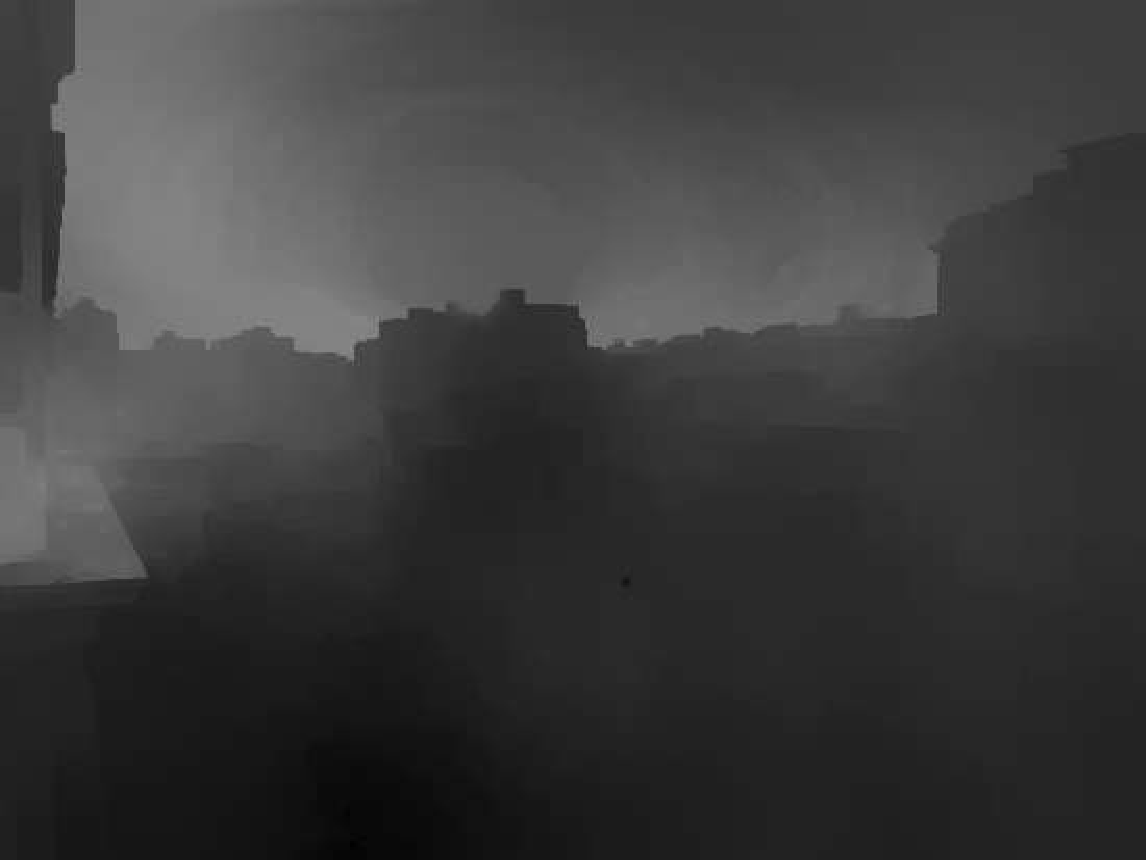}
	\end{minipage}
	\begin{minipage}[b]{.329\linewidth}
		\centering
		\includegraphics[width=1\linewidth]{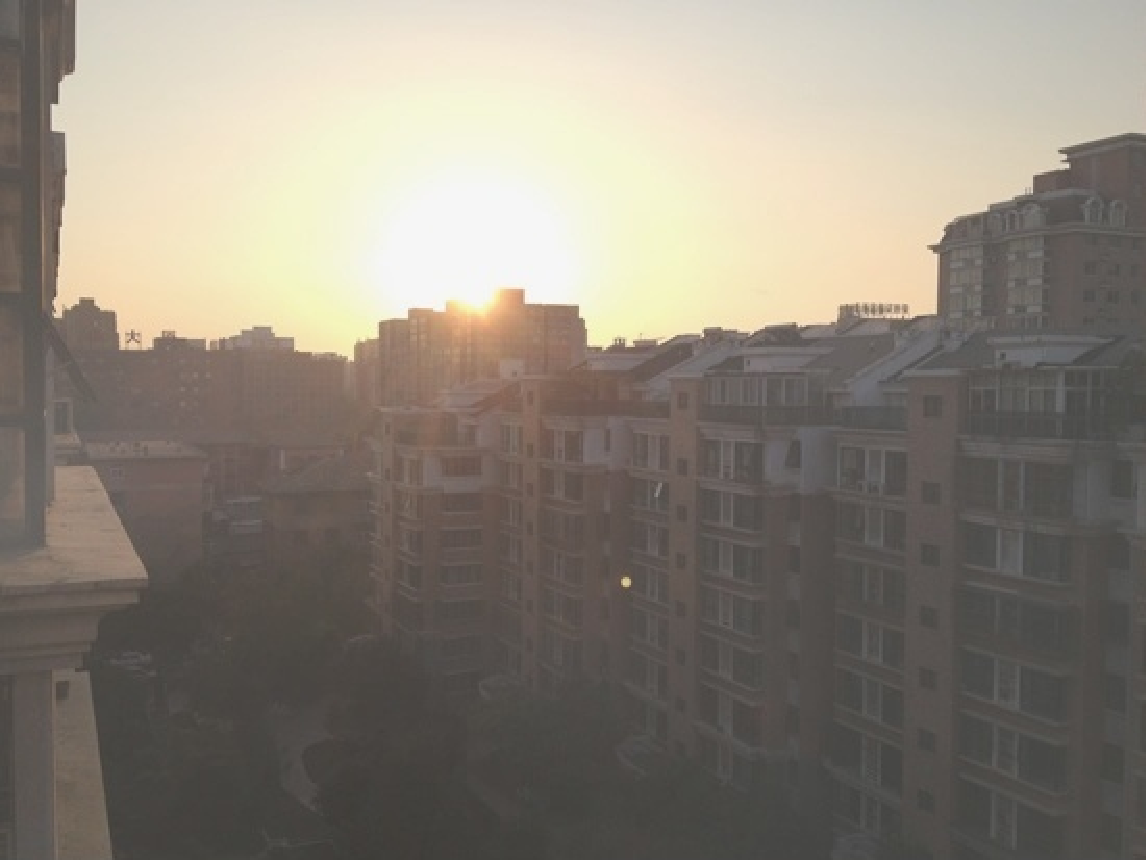}
	\end{minipage}

    \begin{minipage}[b]{.329\linewidth}
		\centering
		\includegraphics[width=1\linewidth]{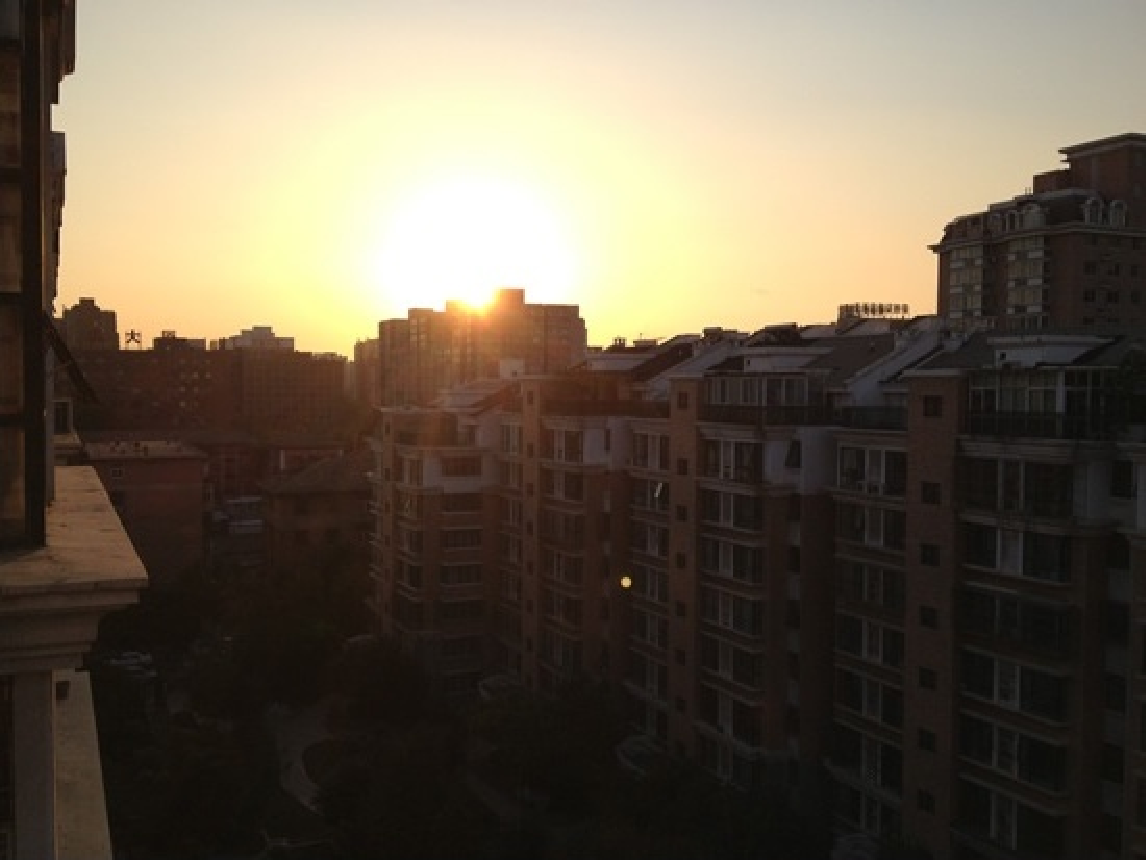}
	\end{minipage}
	\begin{minipage}[b]{.329\linewidth}
		\centering
		\includegraphics[width=1\linewidth]{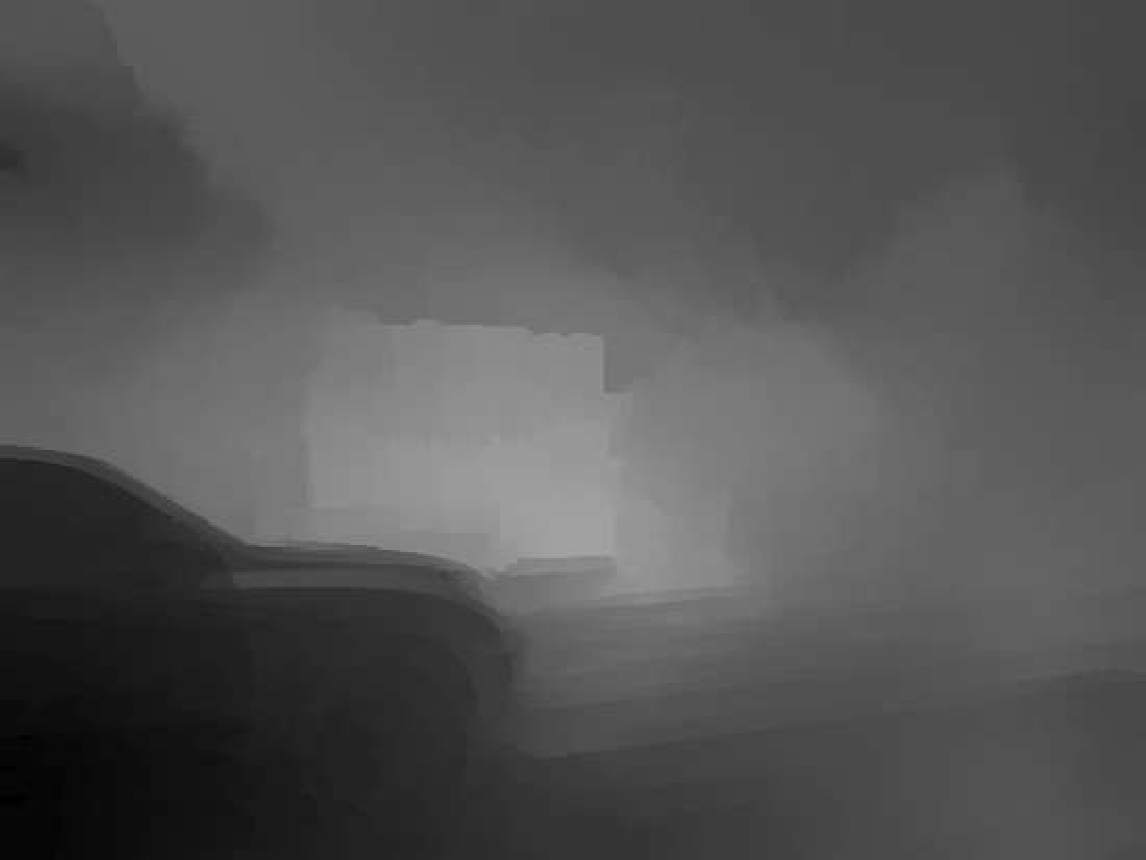}
	\end{minipage}
	\begin{minipage}[b]{.329\linewidth}
		\centering
		\includegraphics[width=1\linewidth]{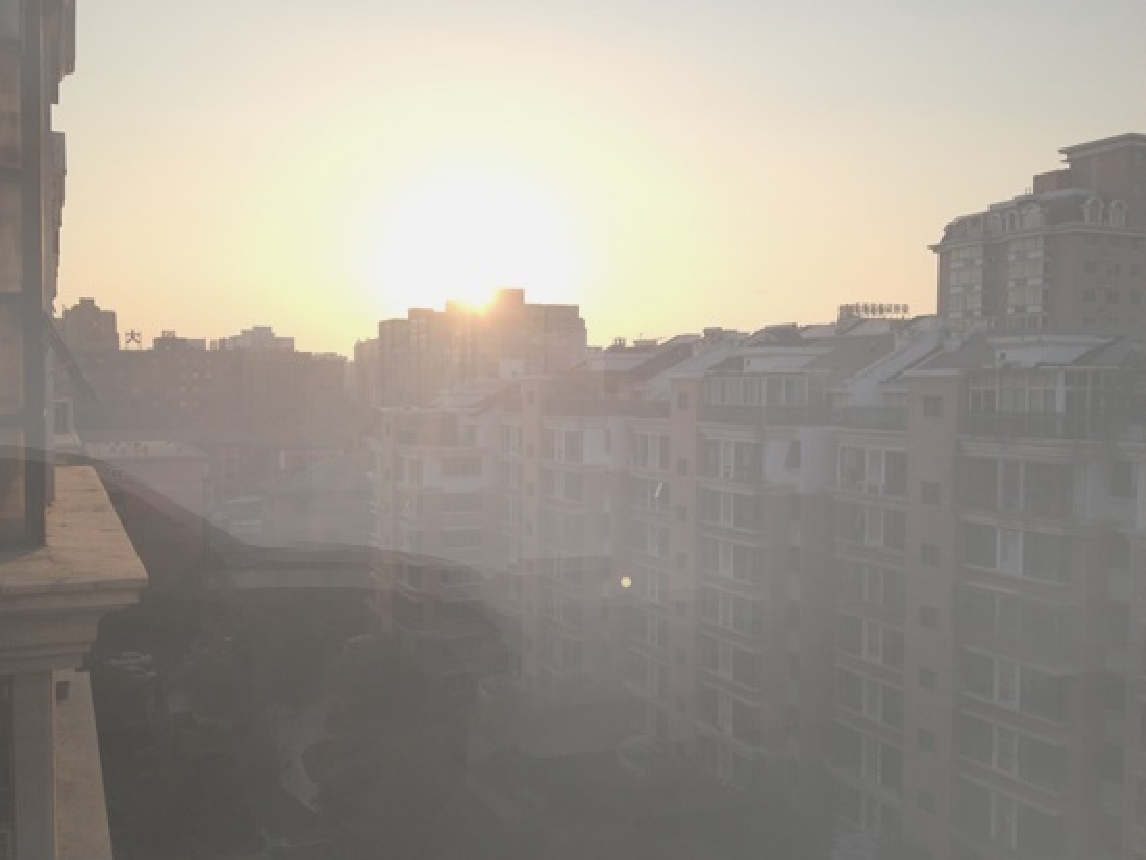}
	\end{minipage}

	
	\caption{Comparisons of previous synthetic method (top line) and ours (bottom line). Left column is the haze-free image, middle column is the depth map to synthesis the hazy image (right column). }
	\label{dahazeimage}
\end{figure}

We propose a novel synthesis method to tackle this problem. Taking inspiration from previous works, we also employ ASM to guide the generation of haze images. Alternatively, as shown in Figure \ref{dahazeimage}, we shuffle the pairs of haze-free images and their corresponding depth maps to decouple the relationship between haze distribution and scene depth. As a result, the vanilla ASM formulation can be modified as follows:

\begin{equation}
\label{asm+}
    I^*=J \cdot t^*+A \cdot(1-t^*),
\end{equation}
\begin{equation}
\label{trans+}
    t^* = e^{- \beta d^*}
\end{equation}

where $t^*$ is the new transmission map updated by $d^*$, a random depth map in the dataset.


The Depth-agnostic Dataset (DA-HAZE) consists of 313,950 image pairs for training and 500 image pairs for testing, which is consistent with OTS. We have also established an evaluation metric termed Discrepancy to assess the dataset's quality. By using different testing sets, denoted as $D_1, D_2, \ldots, D_n$, we compute the variance $Var(R)$ among the dehazing results $R_1, R_2, \ldots, R_n$ obtained from a dehazing method. A lower $Var(R)$ indicates less variation among different testing sets, which demonstrates that the dataset enables the model to perceive real haze distribution.

\textbf{Global Shuffle Strategy.} DA-HAZE decouples the relationship between haze distribution and image depth. However, the haze distribution can dynamically change over time. We propose a Global Shuffle Strategy (GSS) to adapt to this situation.  Specifically, according to Eq.(\ref{asm+}) and Eq.(\ref{trans+}), depth-agnostic hazy images can be generated. Further, for a given haze-free image $J$, there are multiple  $d^*$ associated with it. The number of paired $d^*$ for each image determines the scale of the DA-HAZE. GSS enhances the diversity of the dataset, thereby improving the generalization of the model by allowing it to perceive images with different depth distributions during the training process.   

\subsection{Convolutional Skip Connection }
 U-Net was proposed in the segmentation task, consisting of an encoder path that captures contextual information and a decoder path that performs precise localization. Within this framework, feature fusion methods, such as adding or concatenation, are employed to integrate multi-scale features. For notation simplicity, we use $Z_\theta$ to express the output of a single channel in the feature fusion module, in which $\theta$ represents different operations. The adding operation can be expressed as:
\begin{equation}
\label{add}
    Z_{add} = \sum_{i=1}^{c}\left(X_{i}+Y_{i}\right) * K_{i}=\sum_{i=1}^{c} X_{i} * K_{i}+\sum_{i=1}^{c} Y_{i} * K_{i}
\end{equation}
where $X_i$ and $Y_i$ represent the $i^{th}$ channel of feature maps from skip connection and up-sampling separately. $K$ comprises a set of convolutional kernels, and $K_i$ is the corresponding convolutional kernel of the ith channel. $c$ is the number of channels for each input. The concatenation operation has a similar expression as :

\begin{equation}
\label{concatenate}
    Z_{concat}=\sum_{i=1}^{c} X_{i} * K_{i}+\sum_{i=1}^{c} Y_{i} * K_{i+c}
\end{equation}
Different from adding, the $K$ for concatenation utilizes $2 \cdot c$ convolutional kernels, where the first $c$ kernels are for $X$ and the remains are applied to $Y$. Compared to adding, concatenation has a stronger feature representation ability, thereby achieving better performances.   

\begin{figure}
	
	\begin{minipage}[b]{1\linewidth}
    \centering
	\includegraphics[width=0.95\textwidth]{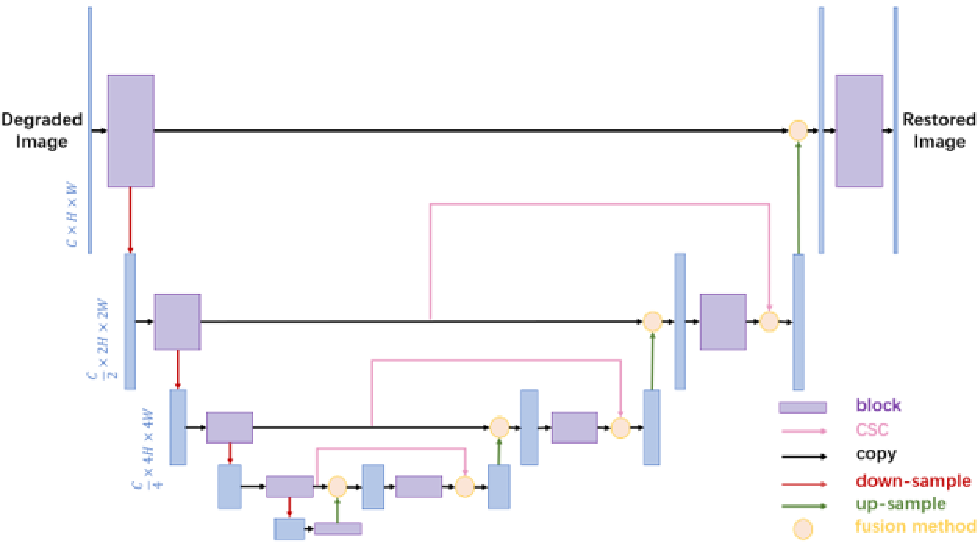}
	\end{minipage}
	\caption{Illustration of Convolutional Skip Connection (CSC). By introducing a single convolution layer (pink line), insufficient feature representations can be mitigated.}
	\label{u1u2r}
\end{figure}

In current dehazing works, adding operation is applied to reduce the input dimension, so that performances and computational costs can be balanced. However, their optimal performances are limited. To this end, we propose a Convolutional Skip Connection module, as shown in Figure \ref{u1u2r}. Specifically, by introducing a single convolution layer, insufficient feature representations can be mitigated. The CSC can be formulated as:   

\begin{equation}
\label{concatenate+}
    Z_{csc}=\sum_{i=1}^{c} X_{i} * K_{i}+\sum_{i=1}^{c} Y_{i} * K_{i}+\sum_{i=1}^{c} Y_{i} * \hat{K}_{i}
\end{equation}
where $\hat{K}$ is the introduced convolution layer that has the same number of channels as $K$. 

Notably, the CSC we proposed can be integrated into any U-Net-based dehazing architecture, with significant improvement in performance but little computational cost.

\subsection{Loss Function}

MSE or L2 loss is the most widely used loss function for single image dehazing. However, some works \cite{lim2017enhanced} pointed out that many image restoration tasks training with L1 loss perform better than L2 loss in terms of PSNR and SSIM metrics. Therefore, we adopt the simple L1 loss. 

\begin{equation}
L_1 = \frac{1}{N}\sum_{i=1}^{n}\vert I_{gt}^i - I_{r}^i \vert
\end{equation}
where \begin{math} I_{gt} \end{math} is the ground truth, \begin{math} I_{r} \end{math} is the restored image.

\section{Experiment}

\subsection{Datasets and Metrics}
\textbf{Datasets.} REalistic Single Image DEhazing (RESIDE) \cite{li2018benchmarking} is a widely used dataset, which consists of Indoor Training Set (ITS, 13,990 image pairs), Outdoor Training Set (OTS, 313,950 image pairs), and Synthetic Objective Testing Set (SOTS, 500 indoor image pairs and 500 outdoor image pairs). We select OTS in the training phase and select SOTS-outdoor in the testing phase for comparison with SOTA methods. DA-SOTS and NH-HAZE, a non-homogeneous realistic dataset, are also used for testing to demonstrate the limitations of OTS. We also train the model on Depth-agnostic Dataset (DH-HAZE) we proposed and test it on SOTS-outdoor, DA-SOTS-outdoor, NH-HAZE \cite{ancuti2020nh}, and O-HAZE \cite{ancuti2018hazeo}.


\textbf{Evaluation Metrics.} Peak signal-to-noise-ratio (PSNR) and structural similarity index (SSIM) \cite{wang2004image}, which are commonly used to measure the image quality among the computer vision community, are utilized for dehazing performance evaluation. 
We also use the \textbf{Discrepancy} described in Sec 3.1 to evaluate the generalization ability of the model.

\subsection{Implementation Details}
The models are optimized by Adam Optimizer. Moreover, Cosine annealing strategy \cite{he2019bag} is adopted, and the
batch size is set to 32. To train the model, we randomly crop patches from the original images with size $256 \times 256$, then two data augmentation techniques are adopted including rotation and vertical or horizontal flip. In the whole training phase, the model is trained for 30 epochs.

\subsection{Comparison with state-of-the-art methods}
NAF-Net \cite{chen2022simple} was proposed as a simple baseline for image restoration. We evaluate its performance on image dehazing by training from scratch on SOTS-outdoor, as shown in Table \ref{nafnet}. NAF-Net outperforms other dehazing methods, with low computational cost. As a complement to NAF-Net, DhazeFormer \cite{song2023vision}, which is a transformer-based architecture is also applied as our baseline. 

\begin{figure}[h]
    \begin{minipage}[b]{.329\linewidth}
		\centering
		\includegraphics[width=1\linewidth]{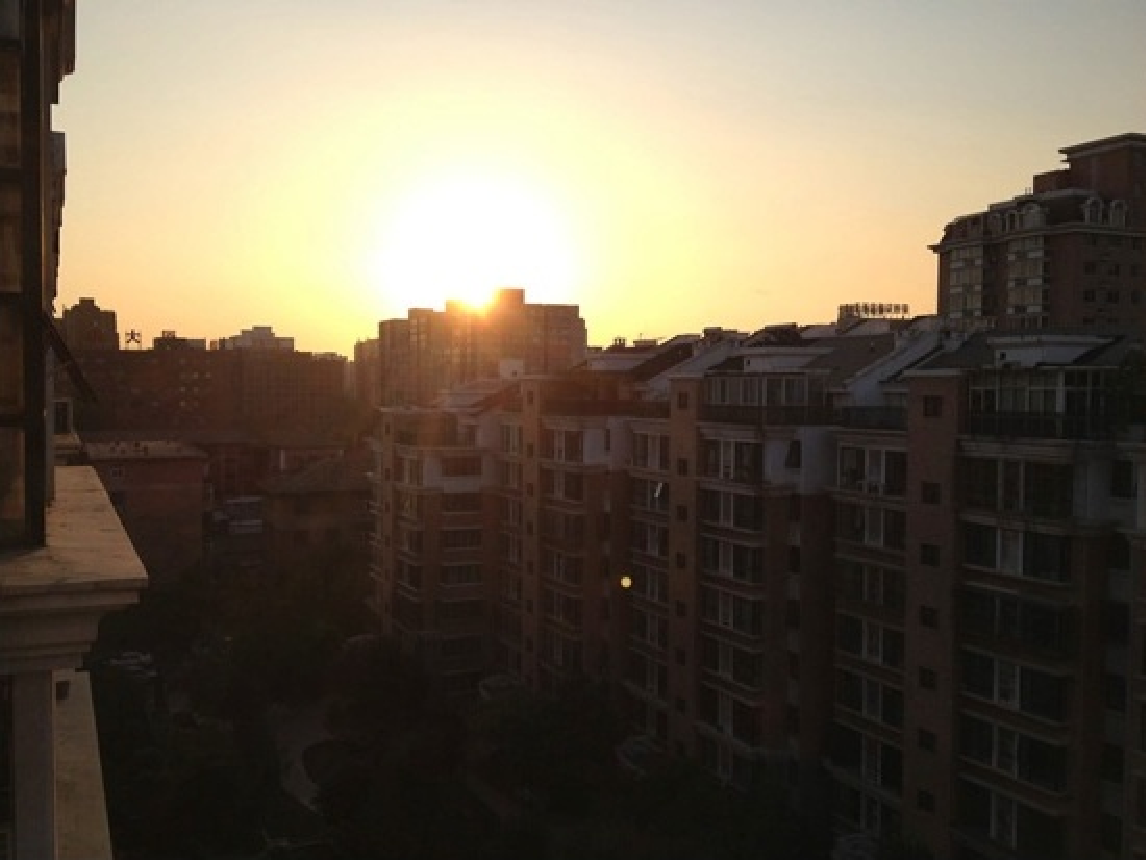}
	\end{minipage}
	\begin{minipage}[b]{.329\linewidth}
		\centering
		\includegraphics[width=1\linewidth]{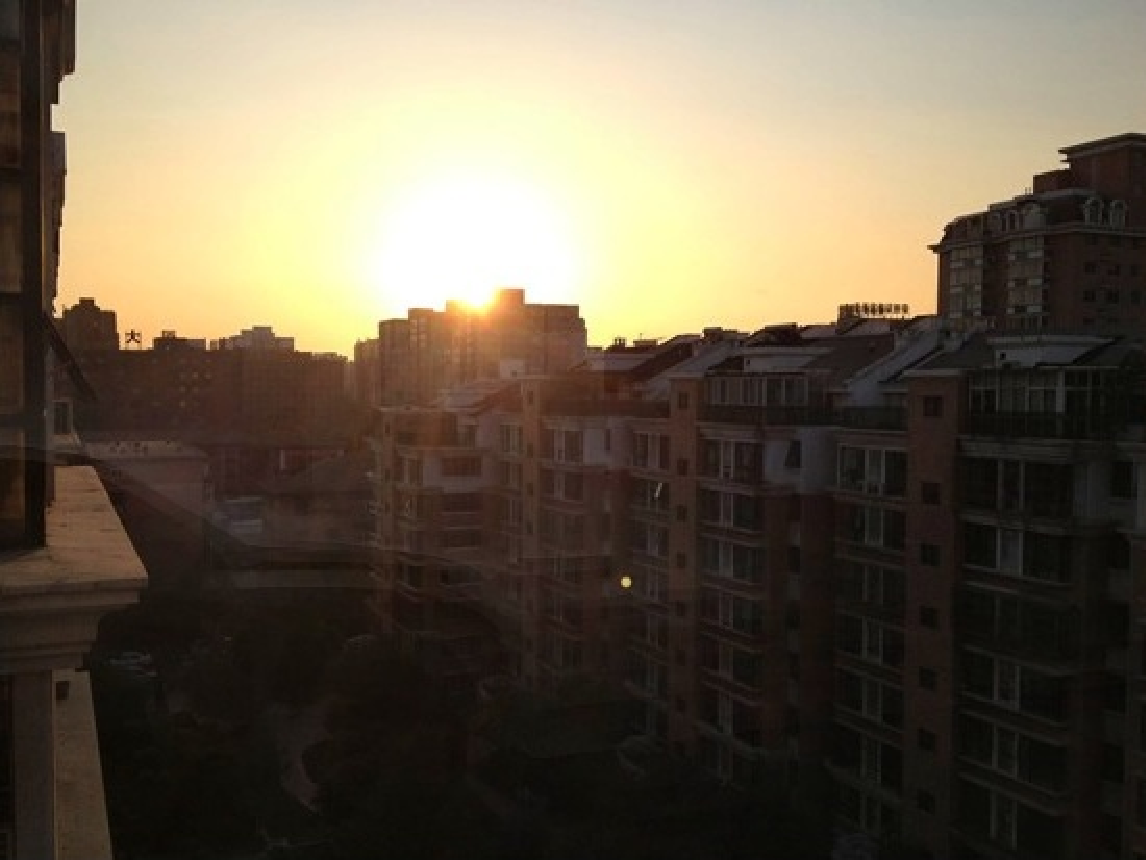}
	\end{minipage}
	\begin{minipage}[b]{.329\linewidth}
		\centering
		\includegraphics[width=1\linewidth]{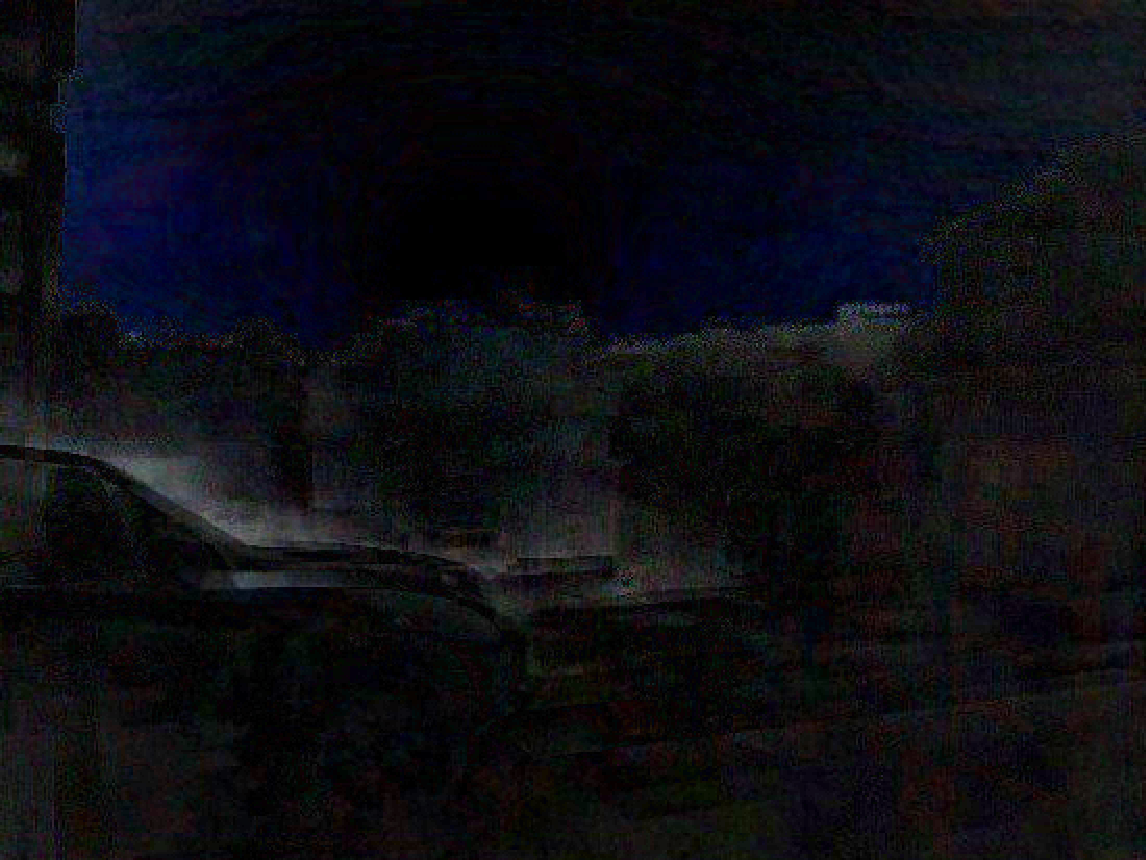}
	\end{minipage}

    \begin{minipage}[b]{.329\linewidth}
		\centering
		\includegraphics[width=1\linewidth]{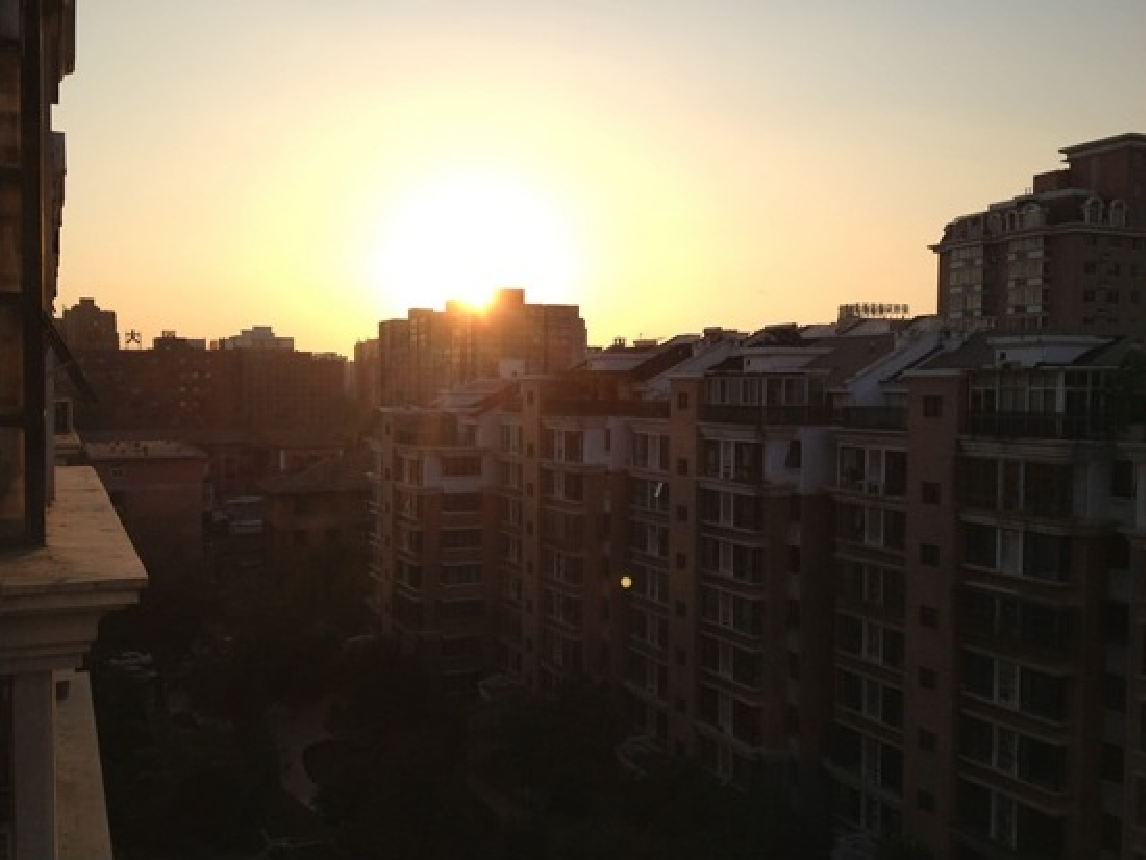}
	\end{minipage}
	\begin{minipage}[b]{.329\linewidth}
		\centering
		\includegraphics[width=1\linewidth]{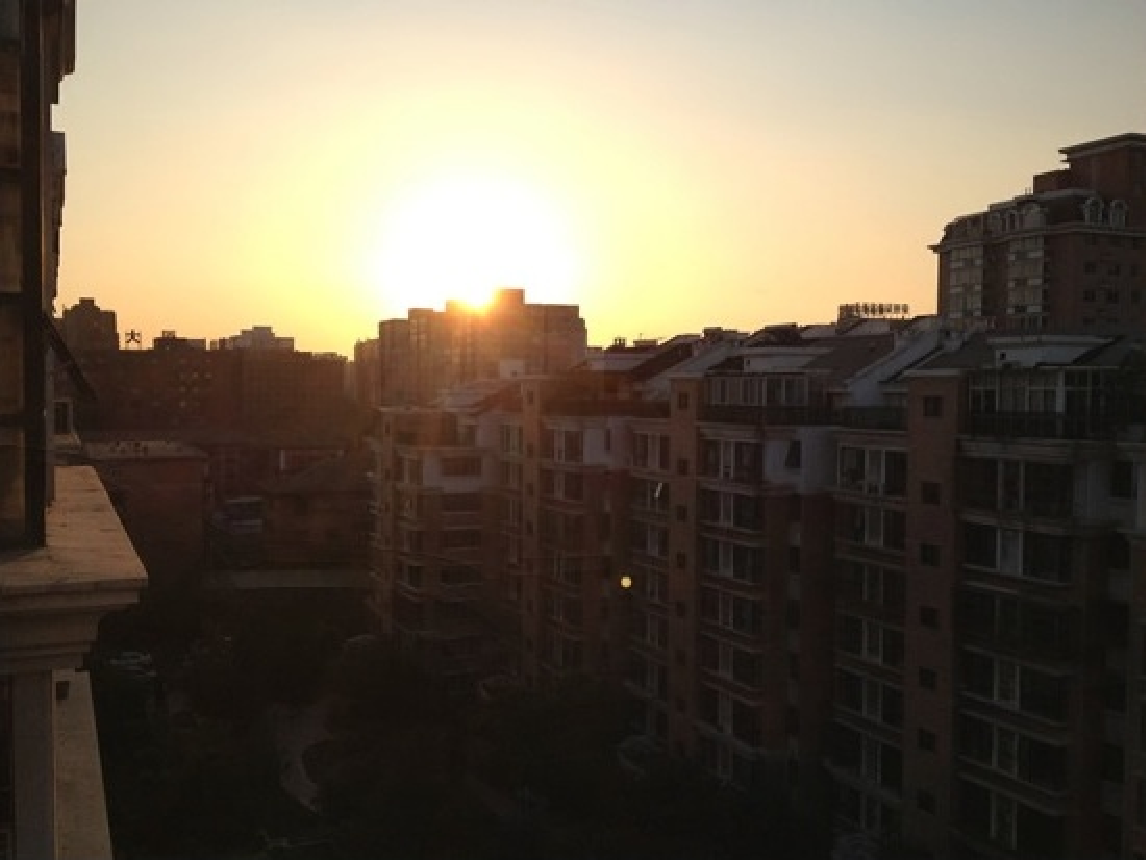}
	\end{minipage}
	\begin{minipage}[b]{.329\linewidth}
		\centering
		\includegraphics[width=1\linewidth]{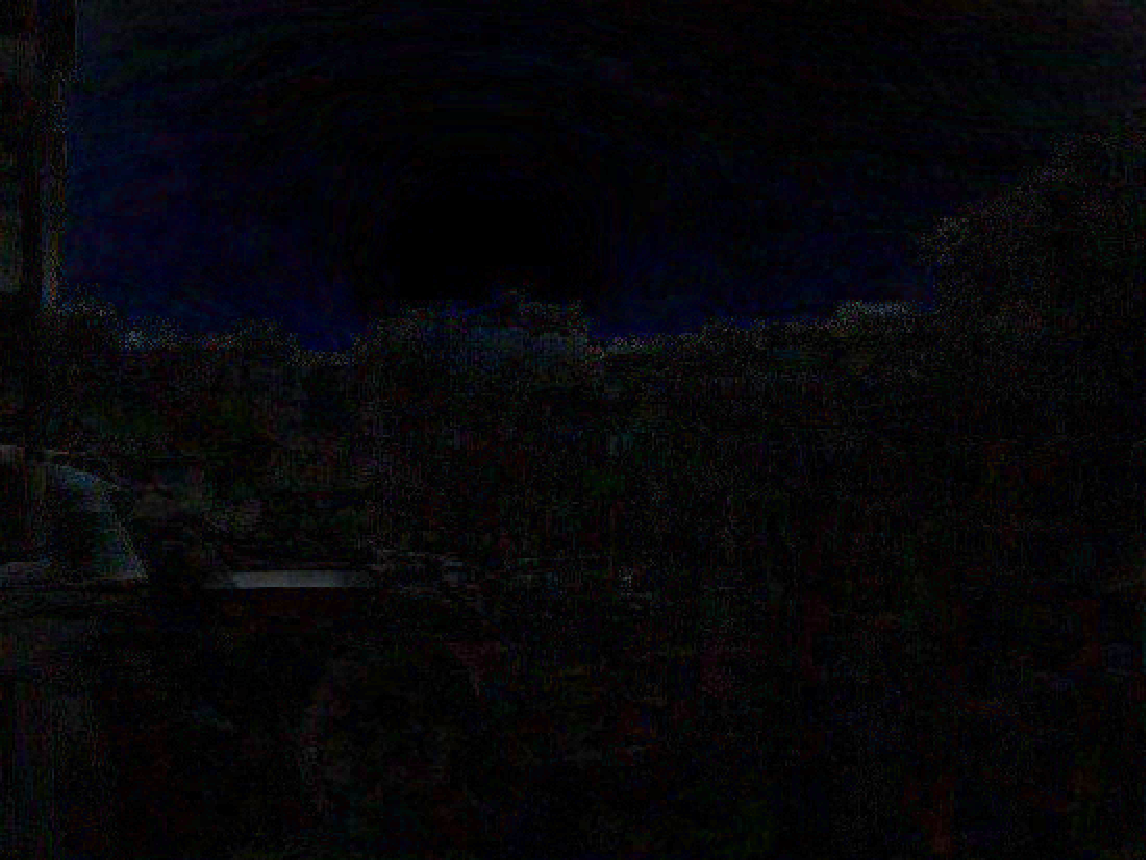}
	\end{minipage}
 
	\caption{Top line shows the restored image of SOTS (left) and DA-HAZE (middle) generated by NAF-Net trained on OTS, and the difference map (right) between them. Bottom line shows the corresponding results by NAF-Net trained on DA-HAZE.}
	\label{vcompare}
\end{figure}

In table \ref{da-haze}, we compare the proposed DA-HAZE with OTS, a classical synthetic dehazing benchmark. Under the same training configurations,  NAF-Net trained on DA-HAZE achieves better performances on real data, with 12.16dB PSNR on NH-HAZE and 16.53dB PSNR on O-HAZE, compared to the one trained on OTS. Similar conclusions can be found in DehazeFormer, which can achieve 0.16dB PSNR and 0.12dB PSNR gains when trained on DA-HAZE. Moreover, we can observe that the model trained on OTS demonstrates a large discrepancy between different hazy distribution test sets (SOTS and DA-SOTS). However, DA-HAZE can significantly mitigate the discrepancy, indicating that it can enable the model to perceive real hazy distribution instead of misleading by depth priors. Some visualization results can be shown in Figure \ref{vcompare}. 

\begin{table}
    \centering
    \caption{Quantitative comparisons of NAF-Net and DehazeFormer trained on OTS and different scaled DA-HAZE, and tested on syntheic and real datasets. Bold indicates the best results.}
    \label{da-haze}
    \resizebox{1\linewidth}{!}{
        \begin{tabular}{c|l|c|cc|cc|c|c|c}
            \hline
            \multirow{2}{*}{Method} &
            \multicolumn{1}{c|}{\multirow{2}{*}{Dataset}} &
            \multirow{2}{*}{Num} &
            \multicolumn{2}{c|}{SOTS-outdoor} &
            \multicolumn{2}{c|}{DA-SOTS-outdoor} &
            discrepancy &
            NH-HAZE &
            O-HAZE \\
            &
            \multicolumn{1}{c|}{} &
            &
            PSNR &
            SSIM &
            PSNR &
            SSIM &
            PSNR &
            PSNR &
            PSNR \\ 
            \hline
            \multirow{4}{*}{NAF-Net}
            & OTS                   & 313950            & \textbf{38.94} & \textbf{0.995} & 32.74 & 0.980 & 9.61 & 11.99 & 16.49 \\
            & DA-HAZE               & 313950            & 34.23          & 0.990          & 35.82 & 0.989 & 0.63 & 12.16 & 16.53 \\
            & DA-HAZE($\times$2)               & $313950 \times 2$ & 35.92          & 0.992          & 37.19 & 0.991 & 0.40 & 12.27 & 16.66 \\
            & DA-HAZE($\times$3)               & $313950 \times 3$ & 36.37 & 0.992  & \textbf{37.52} & \textbf{0.992}& \textbf{0.33}& \textbf{12.33} & \textbf{16.75} \\ 
            \hline
            \multirow{4}{*}{DehazeFormer} 
            & OTS     & 313950            & \textbf{36.44} & \textbf{0.992} & 31.67 & 0.976 & 5.69 & 12.09 & 16.60          \\
            & DA-HAZE & 313950            & 32.30 & 0.978 & 33.52 & 0.984 & 0.37 & 12.25 & 16.72 \\
            & DA-HAZE($\times$2) & $313950 \times 2$ & 33.84 & 0.986 & 34.80 & 0.990 & 0.23 & 12.39 & 16.77 \\
            & DA-HAZE($\times$3) & $313950 \times 3$ & 34.24       & 0.990          & \textbf{35.11} & \textbf{0.991} & \textbf{0.19} & \textbf{12.45} & \textbf{16.89} \\
            \hline
        \end{tabular}
    }
\end{table}

Table \ref{nafnet} shows quantitative evaluation results of DhazeFormer and NAF-Net equipped with our CSC and other state-of-the-art results on SOTS. As we can see, CSC improves the performance on both DhazeFormer and NAF-Net, while introducing minimal computational cost. On SOTS, our NAF-Net-CSC achieves the 38.98dB PSNR, only behind NAF-Net-cat model. However, NAF-Net-CSC contains fewer parameters and flops, implying it can reach a good trade-off between performance and model complexity. 

\begin{table}
    \centering
    \caption{Quantitative comparisons of various dehazing methods on SOTS-outdoor. We report PSNR, SSIM, number of parameters ($\#$ PARAM.) and number of floating-point operations ($\#$ FLOPS). The sign "-" denotes the digit is unavailable. Bold and underlined indicate the best and the second best results respectively.}
    \label{nafnet}
    \begin{tabular}{l|cc|cc}
        \hline
        \multicolumn{1}{c|}{\multirow{2}{*}{Method}} & \multicolumn{2}{c|}{SOTS-outdoor} & \multirow{2}{*}{\# Param. (M)} & \multirow{2}{*}{\# FLOPs (G)} \\
        \multicolumn{1}{c|}{}                      & PSNR  & SSIM  &              &   \\ 
        \hline
        (TPAMI'10) DCP\cite{he2010single}         & 19.14 & 0.861 & -            & -      \\
        (TIP'16) DehazeNet\cite{cai2016dehazenet} & 27.75 & 0.927 & 0.008        & 0.541  \\
        (ECCV'16) MSCNN \cite{ren2016single}      & 22.06 & 0.908 & 0.008        & 0.525  \\
        (ICCV'17) AOD-Net\cite{li2017aod}         & 24.14 & 0.920 & 0.002        & 0.115  \\
        (CVPR'18) GFN\cite{ren2018gated}          & 21.55 & 0.844 & 0.499        & 14.94  \\
        \hline
        (AAAI'20) FFA-Net\cite{qin2020ffa}        & 33.57 & 0.984 & 4.456        & 287.5  \\
        (CVPR'20) MSBDN\cite{dong2020multi}       & 34.81 & 0.986 & 31.35        & 24.44  \\
        (TIP'22) SGID-PFF\cite{bai2022self}       & 30.20 & 0.975 & 13.87        & 152.8  \\
        (AAAI'22) UDN\cite{hong2022uncertainty}   & 34.92 & 0.987 & 4.250        & -      \\
        (ECCV'22) PMDNet\cite{ye2021perceiving}   & 34.74 & 0.985 & 18.90        & -      \\
        (CVPR'22) Dehamer\cite{guo2022image}      & 35.18 & 0.986 & 132.4        & 48.93  \\
        (arxiv'23) DEA-Net\cite{chen2023dea}                                   & 35.97 & 0.989 & 3.653 & 32.23  \\
        \hline
        (TIP'23) DehazeFormer \cite{song2023vision}& 36.44 & 0.992 & 0.686 & 6.658   \\
        DhazeFormer-add   & 36.35 & 0.992  & \textbf{0.683}  & \textbf{6.29}   \\
        DhazeFormer-cat   & 36.63 & 0.992  & 0.797  & 9.26   \\
        \textbf{DhazeFormer-add (w CSC)}   & 36.50 & 0.992  & {\ul 0.686}  & {\ul 6.37}   \\
        \hline
        (ECCV'22) NAF-Net \cite{chen2022simple}             & 38.94   & 0.995    & 29.10          & 16.23  \\
        NAF-Net-cat       & \textbf{39.03} & \textbf{0.995}  & 32.6   & 25.76  \\ 
         \textbf{NAF-Net (w CSC)}       & {\ul 38.98} & {\ul 0.995}  & 29.2   & 16.5   \\
        \hline
    \end{tabular}
\end{table}

\subsection{Abalation studies}
\textbf{The effectiveness of GSS.} To validate the effectiveness of Global Shuffle Strategy (GSS), we change the number of $d^*$ to generate different scaled datasets, by which NAF-Net and Dehaze-Former can be trained. It can be seen from Table \ref{da-haze} that larger-scale datasets can improve the performance of models on real-world data (NH-HAZE and O-HAZE). Meanwhile, as the dataset size increases, the discrepancy between validation sets with different hazy distributions (SOTS and DA-SOTS) also decreases (from 0.63dB PSNR to 0.33dB PSNR on NAF-Net and from 0.37dB PSNR to 0.19dB PSNR on DehazeFormer). The results show that our DA-HAZE with GSS enables the model a strong cross-domain generalization ability.

\textbf{The effectiveness of CSC.} To further validate the superiority of CSC, we replace the fusion methods in DhazeFormer with adding and concatenation operations, termed DhazeFormer-add and DhazeFormer-cat. We incorporate our CSC with DhazeFormer-add and report the results in Table \ref{nafnet}, from which we can see that CSC brings 0.15dB PSNR improvements with introducing minimal computational cost. Compared to DhazeFormer-cat, DhazeFormer-CSC achieves a  better trade-off between performance and computational complexity. Similarly, we equip NAF-Net with CSC, which uses adding operation as a fusion method. It can be found that NAF-Net with CSC achieves competitive results with NAF-Net-cat, which introduces extra parameters. Apart from OTS, CSC also achieves promising results in our proposed DH-HAZE, as shown in Table \ref{csc2}. The model with CSC trained on DH-HAZE shows less discrepancy on different test sets and higher performance on real datasets.  

\begin{table}[]
\centering
\caption{Quantitative comparisons of NAF-Net and DehazeFormer trained on DH-HAZE($\times$3) and tested on synthetic and real datasets. }
\label{csc2}
\resizebox{1\linewidth}{!}{
    \begin{tabular}{l|cc|cc|c|c|c}
    \hline
    \multicolumn{1}{c|}{\multirow{2}{*}{Method}} & \multicolumn{2}{c|}{SOTS-outdoor} & \multicolumn{2}{c|}{DA-STOS-outdoor} & discrepancy & NH-HAZE & O-HAZE \\
    \multicolumn{1}{c|}{} & PSNR  & SSIM  & PSNR  & SSIM  & PSNR & PSNR  & PSNR  \\ \hline
    NAF-Net                    & 36.37 & 0.992 & 37.52 & 0.992 & 0.33 & 12.33 & 16.75 \\ 
    NAF-Net-cat                & 36.68 & 0.992 & 37.63 & 0.992 & 0.23 & 12.69 & 17.12 \\ 
    NAF-Net(w CSC)             & 36.54 & 0.992 & 37.57 & 0.992 & 0.27 & 12.52 & 16.95 \\ \hline
    DehazeFormer-add           & 34.06 & 0.990 & 35.03 & 0.991 & 0.22 & 12.32 & 16.74 \\  
    DehazeFormer-cat           & 34.43 & 0.990 & 35.31 & 0.991 & 0.15 & 12.67 & 17.11 \\ 
    DehazeFormer-add (w CSC)   & 34.32 & 0.990 & 35.18 & 0.991 & 0.18 & 12.50 & 16.94 \\ \hline
    \end{tabular}
}
\end{table}

\section{Conclusion}
In this paper, we propose a novel synthetic method to generate a large-scale dataset, termed DA-HAZE, by which the relationship between haze density and scene depth is decoupled. Meanwhile, a Global Shuffle Strategy (GSS) is proposed to generate different scaled datasets, thereby enhancing the generalization ability of the model. In addition, We propose a Convolutional Skip Connection (CSC) module, allowing for vanilla feature fusion methods to achieve promising results with minimal
costs. Our CSC can be a complement to existing image dehazing methods to enhance their dehazing ability. Extensive experiments demonstrate
that existing methods with CSC can surpass state-of-the-art approaches. Moreover, models trained on DA-HAZE with GSS achieve significant improvements on real-world benchmarks compared to the previous OTS dataset, with less discrepancy
on different distributed validation sets.
%
%
%
\bibliographystyle{splncs04}
\bibliography{mybibliography}
%




\end{document}